\documentclass[conference]{IEEEtran}
\IEEEoverridecommandlockouts
\usepackage{cite}
\usepackage{amsmath,amssymb,amsfonts}
\usepackage{algorithmic}
\usepackage{graphicx}
\usepackage{textcomp}
\usepackage{xcolor}
\usepackage{booktabs}
\usepackage{array}
\usepackage{comment}
\usepackage{url}
\usepackage{booktabs, colortbl, xcolor, array}

\definecolor{bestbg}{HTML}{DBEAFE}
\definecolor{bestfg}{HTML}{1E40AF}
\newcommand{\best}[1]{%
  \cellcolor{bestbg}\textcolor{bestfg}{\textbf{#1}}}

\def\BibTeX{{\rm B\kern-.05em{\sc i\kern-.025em b}\kern-.08em
    T\kern-.1667em\lower.7ex\hbox{E}\kern-.125emX}}

\begin{document}

\title{A 4-Qubit Variational Quantum Classifier for Sequence-Specific Classification of DNMT3A and DNMT3B Flanking Methylation Preferences}

\author{%

    \IEEEauthorblockN{Murat Isik}
    \IEEEauthorblockA{\textit{Purdue University} \\
    West Lafayette, USA \\
    misik@purdue.edu \\
    }
    \and
    \IEEEauthorblockN{Yusuf Moinuddin}
    \IEEEauthorblockA{\textit{Purdue University} \\
    West Lafayette, USA \\
    ymoinudd@purdue.edu \\
    }
    \and
    \IEEEauthorblockN{Humaira Gowher}
    \IEEEauthorblockA{\textit{Purdue University} \\
    West Lafayette, USA \\
    hgowher@purdue.edu}
    
\and
    
    \IEEEauthorblockN{Sabre Kais}
    \IEEEauthorblockA{\textit{NC State University} \\
    Raleigh, USA \\
    skais@ncsu.edu}
        \and

 } 

 \maketitle

\begin{abstract}
DNA methylation is a critical epigenetic modification that regulates gene expression and genome stability, primarily catalyzed by the DNA methyltransferase 3 (DNMT3) family of enzymes. Distinguishing precise sequence-specific flanking preferences between DNMT3A and DNMT3B remains computationally challenging due to the complex and high-dimensional nature of biochemical sequence data. In this study, we propose a novel Quantum Machine Learning (QML) approach to classify DNA methylation preferences, leveraging a Variational Quantum Classifier (VQC). Using experimentally derived deep enzymology data, we constructed a dataset containing 4,096 unique $NNNCGNNN$ motifs. We evaluated a four-qubit VQC using a custom NumPy exact-statevector simulator. The described two-layer circuit contains 24 trainable rotations, eight CNOT gates, and eight input-encoding rotations, totaling 40 gates before measurement. Across five initialization seeds, the revised VQC
evaluation achieved a ROC-AUC of $0.661 \pm 0.009$,
where variability denotes sample standard deviation. The held-out test partition is approximately balanced, containing 200 DNMT3A-labeled motifs and 210
DNMT3B-labeled motifs. An always-DNMT3B predictor achieves 51.2\% accuracy
and a balanced accuracy of 0.500 on this test partition. This proof-of-concept establishes a foundation for utilizing near-term quantum architectures in complex epigenetic classification tasks, paving the way for hybrid quantum-classical pipelines in bioinformatics.
\end{abstract}


\section{Introduction}
DNA methylation is an essential epigenetic mechanism in eukaryotes, governing processes such as cellular differentiation, genomic imprinting, retrotransposon repression, and X-chromosome inactivation. The de novo establishment of genomic methylation landscapes is catalyzed primarily by two closely related paralogous enzymes: DNA methyltransferase 3A (DNMT3A) and DNA methyltransferase 3B (DNMT3B). Despite demonstrating high structural homology in their C-terminal catalytic domains, enzymes fulfill non-redundant biological roles. This divergence is distinctly underscored by distinct phenotypic presentations in knockout target preferences across the mammalian genome. 

DNMT3A and DNMT3B exhibit distinct preferences for the nucleotide sequences flanking CpG sites, leading to enzyme-specific methylation patterns across the genome. These preferences arise from subtle differences in enzyme–DNA interactions and are encoded by complex sequence features that are difficult to resolve with conventional analytical methods. Accurate classification of DNMT3A- and DNMT3B-preferred target sites, therefore, represents an important computational challenge. Machine learning approaches capable of extracting higher-order relationships from sequence data offer a promising strategy for uncovering the sequence determinants governing enzyme-specific methylation activity. Developing predictive computational frameworks capable of learning these high-dimensional sequence relationships is therefore critical for deciphering the rules governing DNMT3A- and DNMT3B-mediated DNA methylation. Such models may ultimately improve our understanding of normal epigenetic programming and the methylation abnormalities associated with human disease.

To unravel these complex biochemical relationships, high-throughput deep enzymology assays have been developed. These assays generate extensive bisulfite sequencing reads that catalog methylation frequencies across thousands of unique variants of the $NNNCGNNN$ motif, providing a rich experimental framework for dissecting the intrinsic targeting preferences of DNMT3A and DNMT3B. Processing and extracting patterns within this high-throughput dataset requires advanced computational methods. Traditionally, classical machine learning paradigms such as Support Vector Machines (SVMs) using string kernels or k-mer features, deep 1D Convolutional Neural Networks (CNNs) optimized for spatial motif identification, and Graph Neural Networks (GNNs) capturing structural nucleotide adjacency have served as the standard approach. More recently, deep Transformer architectures with self-attention heads have demonstrated unmatched capabilities in decoding these sequence arrangements.
These methods have demonstrated considerable success in identifying sequence motifs and higher-order dependencies within genomic datasets. However, as genomic modeling moves beyond pure text representations toward integrating complex multimodal inputs such as Quantum Mechanical/Molecular Mechanical (QM/MM)- derived electronic properties, total self-consistent field (SCF) energy profiles, and structural root-mean-square (RMS) gradients, the underlying feature spaces scale exponentially. This data inflation introduces the curse of dimensionality, taxing classical regularization bounds \cite{schuld2020circuit, farhi2018classification, du2020expressive}. Therefore, the search for alternative computational frameworks capable of efficiently representing complex sequence relationships remains an active area of investigation.

Quantum Machine Learning (QML) has emerged as a promising paradigm for pattern recognition and classification tasks. It offers an innovative pathway to navigate these multidimensional biological features \cite{isik2025multimodal, singh2026quantum, bhatia2023quantum, saggi2024mqml}. By encoding classical data strings directly into the exponentially large Hilbert spaces accessible via parameterized quantum circuits, quantum architectures can exploit state-space interference patterns. Parameterized quantum circuits provide nonlinear feature
representations, but their predictive value must be
established through controlled classical comparisons \cite{havlivcek2019supervised, monard1999combined}. 

Within the broader DNMT family, DNMT1 serves as the 
maintenance methyltransferase responsible for copying 
established methylation patterns during DNA replication, 
while DNMT3A and DNMT3B act as de novo methyltransferases 
that establish new methylation landscapes~\cite{zhong2014molecular}. 
Although these enzymes share substantial structural similarity, they exhibit distinct preferences for nucleotide sequences flanking CpG sites, resulting in characteristic methylation signatures that can be measured experimentally using deep enzymology assays. ~\cite{okano1999dna}.

We investigate a four-qubit variational quantum classifier
for distinguishing DNMT3A and DNMT3B sequence preferences.
The revised evaluation uses noiseless statevector simulation
and compares the VQC with classical models using identical
features, data partitions, and implement rigorous validation strategies to ensure that model predictions reflect biologically meaningful sequence determinants rather than dataset-specific artifacts. \cite{preskill2018quantum, bharti2022noisy}. The main contributions are:
\begin{itemize}
    \item A dataset containing 4,096 unique $NNNCGNNN$
    motifs, partitioned without motif overlap.
    \item A compact VQC evaluated using a fixed
    training, validation, and test protocol.
    \item Classical comparisons using identical inputs,
    with initialization variability reported across seeds.

    \end{itemize}

\section{Related Work}
\subsection{Classical Sequence Analysis in Epigenetics}
The characterization of enzyme-substrate specificity for mammalian methyltransferases has historically relied on position weight matrices (PWMs) and consensus motif identification. While PWMs provide highly interpretable representations of single-nucleotide independent contributions, they fail to capture multi-nucleotide epistatic dependencies across flanking positions. To address these limitations, deep learning frameworks have been widely adopted. CNNs that use parallelized one-dimensional convolutional kernels effectively capture local sequence dependencies and spatial configurations, thereby simulating protein-DNA binding affinities. Furthermore, the incorporation of Transformer networks that utilize multi-head scaled dot-product attention has enabled the resolution of long-range nucleotide correlations. Despite these classical breakthroughs, resolving highly correlated physical properties across closely related enzyme classes remains a substantial bottleneck, requiring high parameter counts and increasing the risk of overfitting \cite{lo2016deep}.

\subsection{Quantum Machine Learning for Bioinformatics}
The intersection of QML and computational biology is an expanding domain of research, motivated by the mathematical alignment between quantum state spaces and the combinatorial complexity of biological sequences \cite{schuld2019quantum, chakraborty2023quantum}. Early implementations of quantum-assisted pipelines focused primarily on protein folding optimization using Quantum Annealing frameworks. Recently, gate-based quantum machine learning models, particularly VQCs and Quantum Support Vector Machines (QSVMs), have been introduced to classify biological text data, splice-site boundaries, and transcriptomic profiles \cite{panizza2025quantum, torlai2018neural, huembeli2021characterizing}. Parameterized quantum circuits operate as quantum neural networks, adjusting internal rotation parameters via classical optimizers. While these models show theoretical promise, applying them to highly imbalanced, experimentally derived deep enzymology data remains an open challenge that requires rigorous validation to verify genuine quantum advantages \cite{lloyd2018quantum, gelman2024survey, wierichs2022general}.

\section{Methodology}
\subsection{Data Engineering and Pipeline Architecture}
The foundational data utilized in this study are derived from deep enzymology bisulfite sequencing assays designed to track competitive methylation mechanics. A key limitation in conventional genomic classification is the reliance on file-membership labeling, where any sequence read found within an isolated enzyme file is assumed to represent a clean preference signal.

\begin{figure}[htbp]
\centerline{\includegraphics[width=0.48\textwidth]{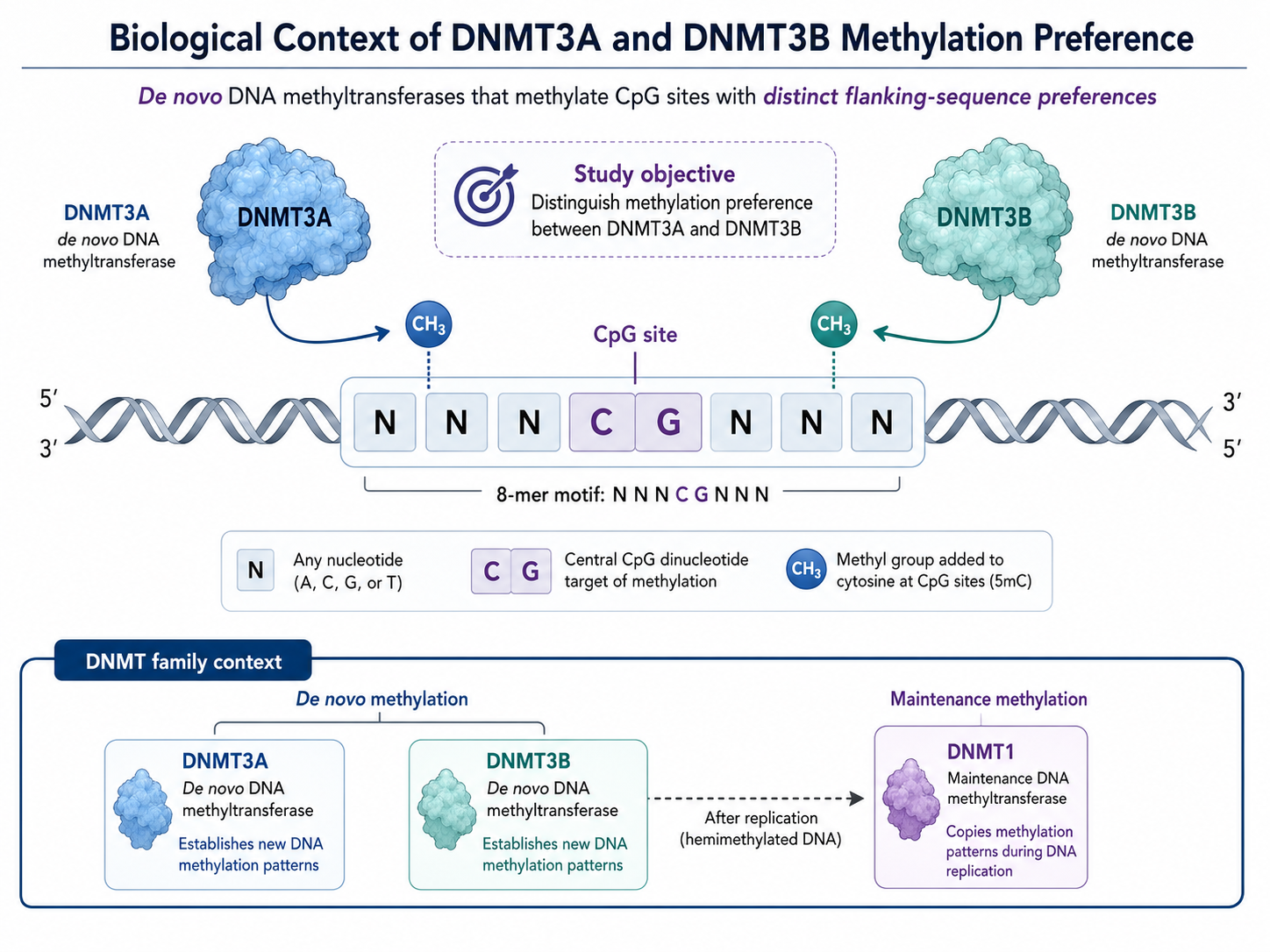}}
\caption{Biological context of DNMT3A and DNMT3B 
methylation preferences. The schematic details the 
target 8-mer motif ($NNNCGNNN$) alongside the functional 
differentiation within the DNMT family, contrasting 
de novo methylation by DNMT3A and DNMT3B with 
maintenance methylation by DNMT1. Methyl group addition 
to cytosine at CpG sites produces 5-methylcytosine (5mC).}
\label{fig:biological_context}
\end{figure}

We describe the dataset, sequence representation,
classifier, and evaluation protocol.
Fig.~\ref{fig:biological_context} illustrates the
biological classification task.

The revised analysis uses Supplementary Data~1 of Gao et al.
Let $A(m)$ and $B(m)$ denote the reported relative activities
of DNMT3A and DNMT3B for motif $m$. Labels are defined as
\[
y_m =
\begin{cases}
1, & B(m)>A(m),\\
0, & B(m)\leq A(m).
\end{cases}
\]
Class 1 denotes DNMT3B-favored motifs; ties are assigned
to class 0 under this rule. The labels describe relative
activity under the source experimental conditions.

To suppress stochastic sequencing noise and eliminate low-confidence entries, a minimum-count filter constraint was enforced such that:

$$\sum R_{A}(m) + \sum R_{B}(m) \ge 5$$

\begin{figure}[htbp]
\centerline{\includegraphics[width=0.48\textwidth]{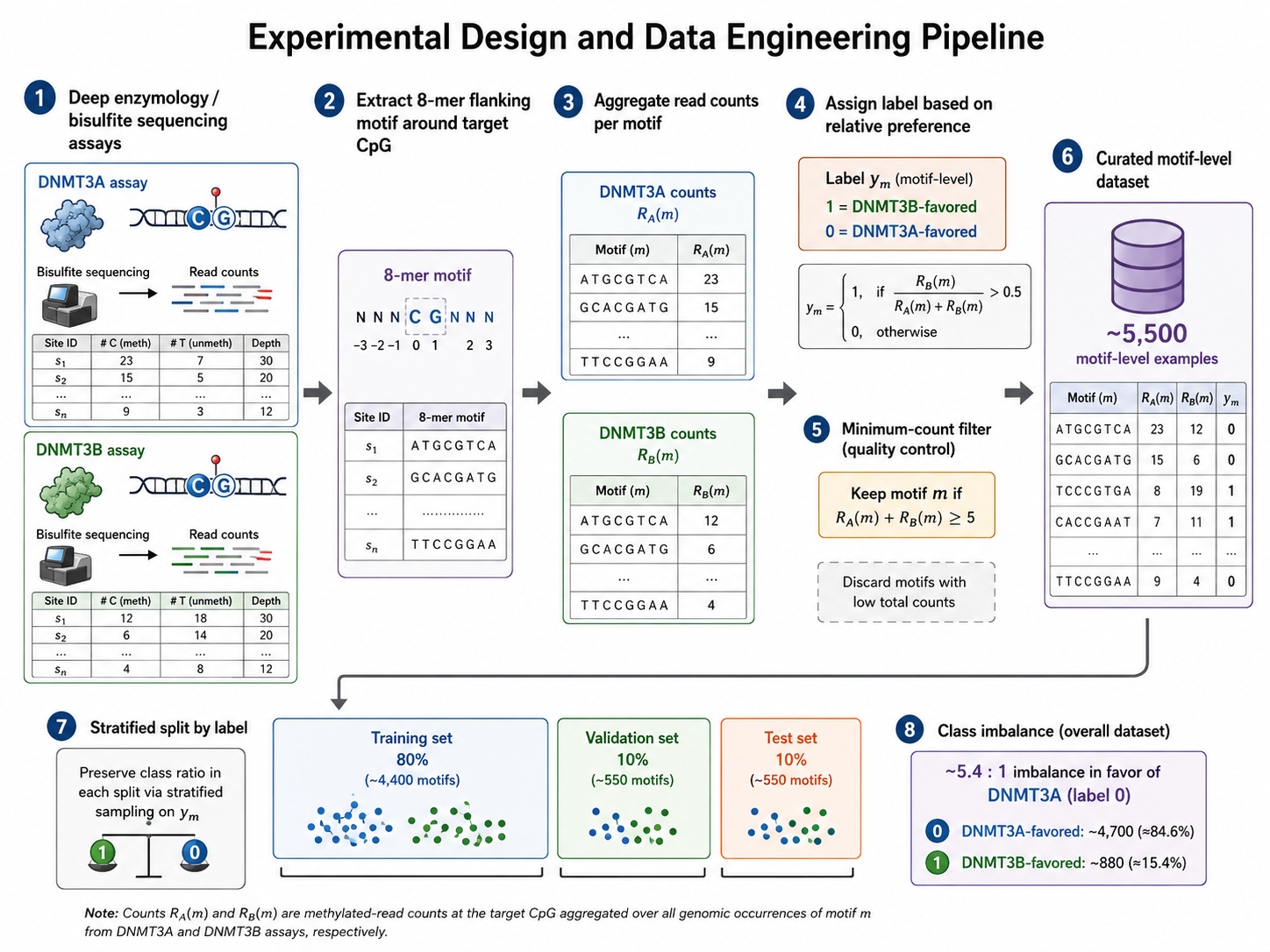}}
\caption{Dataset preparation and evaluation workflow for
DNMT3A/DNMT3B preference classification. The revised dataset
contains 4,096 unique $NNNCGNNN$ motifs, with one entry per
motif. A fixed stratified split assigns 3,276 motifs to
training, 410 to validation, and 410 to testing, with no
motif overlap between partitions. Validation data are used
for model and checkpoint selection; test data are reserved
for final evaluation.}
\label{fig:pipeline_architecture}
\end{figure}

The revised dataset contains 4,096 unique $NNNCGNNN$ motifs,
with one entry per motif. Each motif contains six variable
nucleotide positions and a fixed central CpG dinucleotide,
permitting $4^6 = 4096$ possible unique sequences. We used a fixed stratified 80/10/10 split with random
state 42, yielding 3,276 training motifs, 410 validation
motifs, and 410 test motifs. The partitions contain no
overlapping motifs. Model and checkpoint selection used
the validation partition, while the test partition was
reserved for final evaluation. Fig.~\ref{fig:pipeline_architecture} summarizes the dataset
preparation and evaluation workflow.

\subsection{Feature Mapping and State Initialization}
The translation of classical discrete genomic text strings into coherent quantum states requires structured feature mapping. 
Each nucleotide within the 8-mer motif $m = (n_1, n_2, \dots, n_8)$ 
is first converted into a dense classical representation using 
a fixed ordinal encoding: A $\mapsto$ 0.0, C $\mapsto$ 1.0, 
G $\mapsto$ 2.0, T $\mapsto$ 3.0. To achieve a compact 
configuration compatible with near-term 4-qubit constraints, 
pairs of adjacent nucleotides are condensed and linearly 
normalized into numerical vectors, mapping the input features 
to a continuous domain $x \in [-\pi, \pi]^4$, as illustrated 
in Fig.~\ref{fig:vqc_architecture}.

The numerical vector $x$ is embedded into the quantum state 
space via a parameterized feature map unitary operator, 
$\mathcal{U}_{\Phi}(x)$. For a 4-qubit initialization register, the classical vector maps onto the zero-state base configuration:

$$|\psi_x\rangle = \mathcal{U}_{\Phi}(x) |0\rangle^{\otimes 4}$$

Each angle feature is encoded using an $R_Y$ rotation
followed by an $R_Z$ rotation on the corresponding qubit,
starting from $|0\rangle^{\otimes 4}$. This encoding
contributes eight single-qubit rotations.

\subsection{Variational Quantum Circuit Topology}
The VQC uses four qubits and two variational layers.
Each layer contains four trainable $R_Y$, four trainable
$R_X$, and four trainable $R_Z$ rotations, together with
a ring of four CNOT gates. The variational ansatz therefore
contains 24 trainable rotation parameters and eight CNOT
gates. With eight input-encoding rotations, the circuit
contains 40 gates before measurement, excluding any
compiler decomposition.

The entire variational ansatz can be mathematically 
represented as a product of parameterized unitary 
transformations across $L$ layers:

$$\mathcal{W}(\theta) = \prod_{l=1}^{L} U_l(\theta_l)$$

A scalar circuit observable $s(x;\theta)$ is passed to
an affine classical readout:
\[
\ell = a\,s(x;\theta) + b,
\qquad p = \sigma(\ell).
\]
The readout contributes two trainable parameters, $a$ and
$b$, giving 26 parameters total showed in Table~\ref{tab:hardware_specs}.

\subsection{Loss Function and Optimization}
\begin{figure}[htbp]
\centerline{\includegraphics[width=0.48\textwidth]{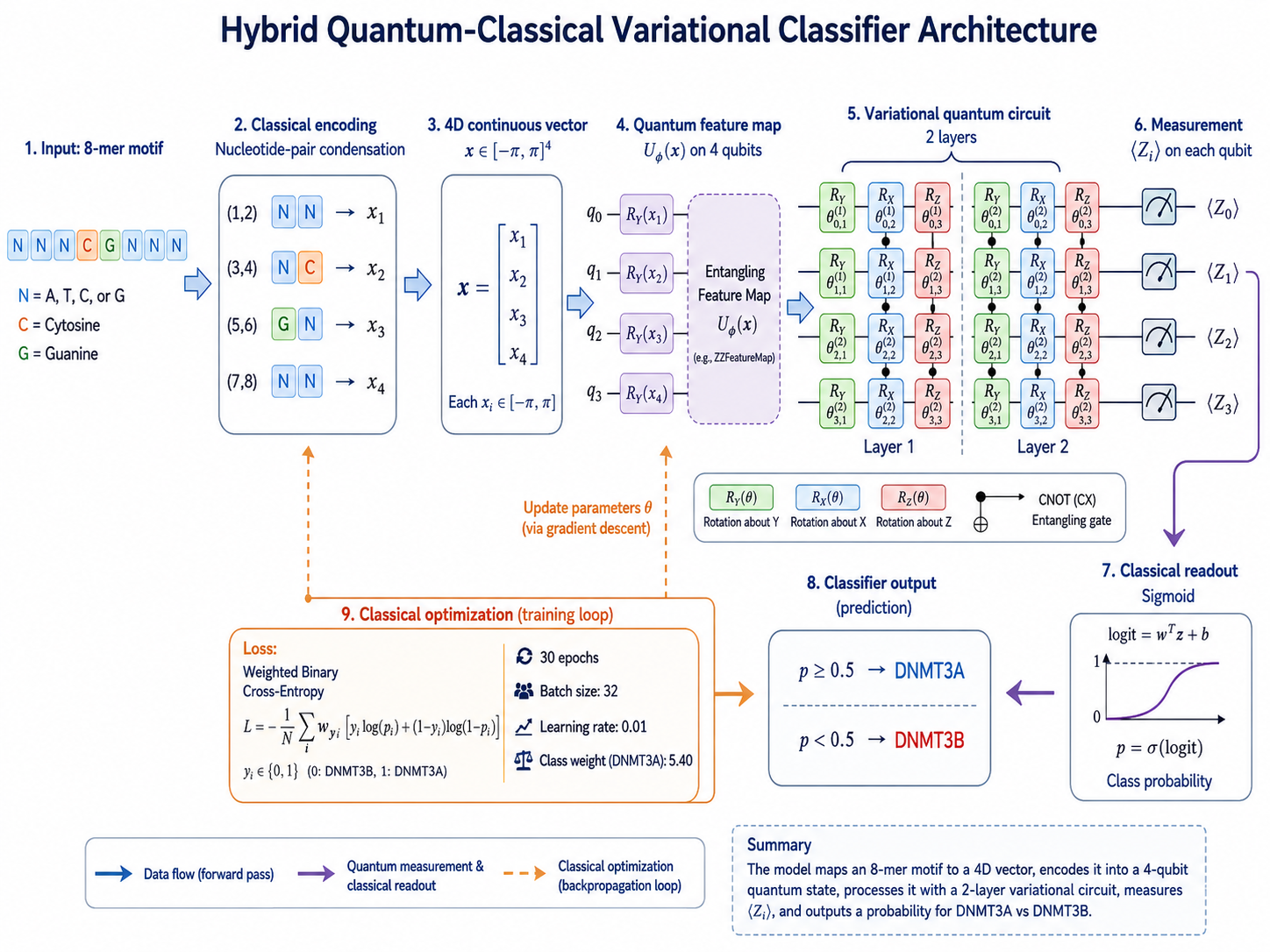}}
\caption{Hybrid Quantum-Classical Variational Classifier 
Architecture. The diagram illustrates the complete forward 
data pass from an 8-mer genomic motif input through 
nucleotide-pair condensation and ordinal encoding 
(A$\mapsto$0, C$\mapsto$1, G$\mapsto$2, T$\mapsto$3), 
4D feature mapping to $x \in [-\pi, \pi]^4$, 
2-layer parameterized quantum circuit execution, 
4-qubit Pauli-Z measurement, linear readout, sigmoid 
activation, and weighted binary cross-entropy 
backpropagation. The sigmoid output represents the DNMT3B class score:
$p \geq 0.5$ predicts DNMT3B (label 1), while
$p < 0.5$ predicts DNMT3A (label 0).}
\label{fig:vqc_architecture}
\end{figure}

We train the classifier using weighted binary cross-entropy:
\[
\mathcal{L}
= -\frac{1}{N}\sum_{i=1}^{N}
\left[
w y_i \log p_i +
(1-y_i)\log(1-p_i)
\right],
\]
where $p_i$ is the sigmoid score for DNMT3B (class 1).
The positive-class weight is calculated exclusively from
the training partition as
\[
w = \frac{n_{0,\mathrm{train}}}{n_{1,\mathrm{train}}},
\]
where $n_{0,\mathrm{train}}$ and $n_{1,\mathrm{train}}$
denote the DNMT3A and DNMT3B training counts, respectively. Throughout this study, class 0 denotes DNMT3A and class 1
denotes DNMT3B. The sigmoid output $p$ is the model score
for the positive class, DNMT3B. Predictions are obtained
using a fixed threshold of 0.5:
\[
\hat{y} =
\begin{cases}
1 \; (\mathrm{DNMT3B}), & p \geq 0.5,\\
0 \; (\mathrm{DNMT3A}), & p < 0.5.
\end{cases}
\]
. For each initialization seed, the checkpoint is selected
using validation average precision. The test partition is
reserved for evaluation after checkpoint selection. The 
complete end-to-end hybrid quantum-classical 
architecture is systematically detailed in 
Fig.~\ref{fig:vqc_architecture}.

\subsection{Classical Baselines}
We compared the four-qubit VQC with logistic regression,
a support vector machine with a radial basis function
kernel (RBF-SVM), a random forest containing 500 trees,
and two compact multilayer perceptrons (MLPs) containing
25 and 31 trainable parameters. These MLPs bracket the
reported 26-parameter VQC in parameter count, although
similar parameter counts do not imply equivalent model
capacity or computational cost. For the primary comparison, all models received the same
four ordered-pair angle features and used the same fixed,
motif-disjoint training, validation, and test partitions.
Class weights were derived from training labels only.
Hyperparameter and checkpoint selection used validation
average precision; the test partition was reserved for
final evaluation. A separate sensitivity analysis used a 24-dimensional
one-hot representation of the six variable nucleotide
positions. We report this analysis separately because it
changes the input representation and therefore does not
constitute an identical-input comparison with the VQC.

\section{Experimental Results}

\subsection{Baseline Classification Metrics}
The revised evaluation uses a held-out test partition containing
410 unique motifs: 200 assigned to DNMT3A (class 0) and 210
assigned to DNMT3B (class 1). Model and checkpoint selection
uses validation average precision, with the test partition
reserved for final evaluation.

Across five random seeds (0--4), the exact-statevector VQC
achieved a mean ROC-AUC of $0.661 \pm 0.009$, where the
reported variability is the sample standard deviation across
seeds. This summarizes sensitivity to random initialization
on the fixed data split; it is not a confidence interval
or evidence of generalization to independent experiments.

The test partition is approximately balanced. An always-DNMT3B
predictor achieves an accuracy of $210/410 \approx 0.512$
and a balanced accuracy of 0.500. The positive-class prevalence
reference for precision-recall evaluation is also
$210/410 \approx 0.512$.

\begin{figure}[htbp]
  \centering
  \includegraphics[width=\columnwidth]{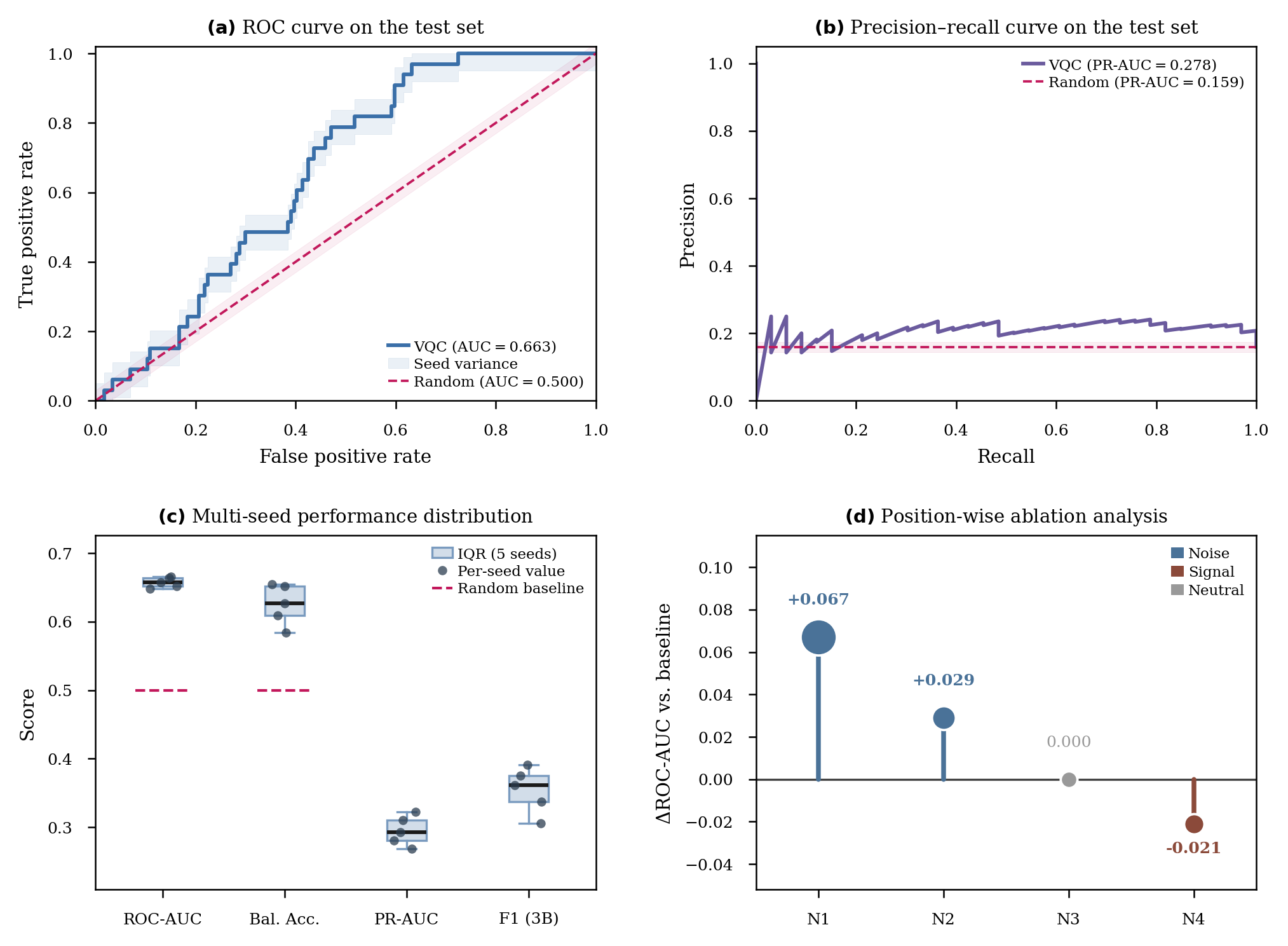}
  \caption{Earlier evaluation plots, retained separately
from the revised analysis.
\textbf{(a)} ROC curve.
\textbf{(b)} Precision--recall curve.
\textbf{(c)} Multi-seed performance distributions.
\textbf{(d)} Changes in ROC-AUC under feature masking.
Embedded values have not been reconciled with the
revised test partition.}
  \label{fig:combined_results}
\end{figure}

Fig.~\ref{fig:combined_results} retains the earlier
evaluation plots, which have not been reconciled with
the revised 410-motif test partition. The panels show
(a) the ROC curve, (b) the precision--recall curve,
(c) multi-seed performance distributions, and
(d) feature-ablation results. The displayed
precision--recall reference of 0.159 differs from the
revised test prevalence of approximately 0.512.
These plots do not substantiate the revised results.

Under the identical-input comparison, the RBF-SVM achieved
a test ROC-AUC of 0.856, while the random forest achieved
$0.896 \pm 0.002$, compared with $0.661 \pm 0.009$ for
the VQC. The reported variability denotes sample standard
deviation across five seeds for the random forest and VQC.
These results favor the classical baselines and provide
no evidence of a predictive advantage for the VQC under
the evaluated configuration.

\subsection{Simulated Hardware Profiling}
To understand the operational overhead required for deployment onto
physical Quantum Processing Units (QPUs), we systematically tracked
performance metrics directly from the Qiskit Aer simulation
environment \cite{b10}. This step is critical for evaluating whether near-term
quantum architectures can handle high-throughput bioinformatics
workflows.
The circuit achieved an average execution latency of 1.12 ms per
circuit. This latency profile translates to a sustained processing
throughput of 889.09 circuits per second under a standardized
resource allocation. Furthermore, the maximum memory footprint
remained highly optimized, demanding a peak classical memory overhead
of only 0.01 MB. This efficient profile underscores the advantages
of compact VQC architectures compared to high-parameter classical
deep learning models. Primary training and inference use a custom NumPy
exact-statevector simulator. No gate noise, readout noise,
or decoherence model is included. A separate evaluation
uses 256 measurement shots per test motif to assess
finite-sampling effects. The hardware parameters are categorized in
Table~\ref{tab:hardware_specs}. The earlier profiling table reports a 36-gate circuit,
whereas the revised circuit description contains
40 gates. The retained timing measurements therefore
cannot be attributed to the revised circuit without
verification.

\begin{table}[htbp]
\caption{Earlier reported circuit specifications and
Qiskit Aer profiling measurements.}
\label{tab:hardware_specs}
\begin{center}
\begin{tabular}{ll}
\toprule
\textbf{Hardware/Circuit Parameter} & \textbf{Observed Metric Value} \\
\midrule
Register Allocation        & 4 Operational Qubits \\
Evaluation Resolution      & 256 Shots \\
Total Structural Gate Count & 36 Gates \\
Parameterized Rotation Gates & 24 (\texttt{ry}: 8, \texttt{rx}: 8, \texttt{rz}: 8) \\
Entangling Operations      & 8 CNOT (\texttt{cx}) Gates \\
Measurement Operators      & 4 Pauli-Z ($\langle \mathcal{Z}_i \rangle$) Operators \\
Trainable Parameters       & 26 (24 VQC + 2 Linear) \\
Average Execution Latency  & 1.12 ms per circuit \\
Simulation Throughput Capacity & 889.09 circuits/sec \\
Peak Resource Memory Allocation & 0.01 MB \\
\bottomrule
\end{tabular}
\end{center}
\end{table}

\subsection{Multi-Seed Statistical Robustness}
To rule out performance anomalies arising from favorable random weight
initializations, the full VQC training and evaluation pipeline was
executed across five independent random seeds (seeds 0--4), with each
run employing independent parameter initialization and best-checkpoint
selection. Results are reported as mean $\pm$ standard deviation
across seeds.

The revised analysis reports a ROC-AUC of
$0.661 \pm 0.009$ across five initialization seeds.
This summarizes initialization variability on a fixed
data split and does not establish generalization to
independent experiments. Table~\ref{tab:multi_seed}
contains the earlier per-seed results and does not
substantiate the revised summary.

\begin{table}[htbp]
\caption{Earlier per-seed evaluation results, retained
separately from the revised analysis.}
\label{tab:multi_seed}
\centering
\resizebox{\columnwidth}{!}{%
\setlength{\tabcolsep}{6pt}
\renewcommand{\arraystretch}{1.25}
\begin{tabular}{c c c c c c c}
\toprule
\textbf{Seed} &
\textbf{\shortstack{Best\\Epoch}} &
\textbf{Accuracy} &
\textbf{\shortstack{Bal.\\Acc.}} &
\textbf{ROC-AUC} &
\textbf{PR-AUC} &
\textbf{\shortstack{F1\\(DNMT3B)}} \\
\midrule
0 & 25 & 0.671 & 0.584 & 0.648 & 0.269 & 0.306 \\
1 & 15 & 0.662 & 0.652 & \best{0.666} & \best{0.322} & \best{0.375} \\
2 & 13 & 0.729 & \best{0.655} & 0.657 & 0.310 & 0.391 \\
3 & 14 & 0.715 & 0.609 & 0.664 & 0.281 & 0.337 \\
4 &  7 & \best{0.744} & 0.627 & 0.652 & 0.293 & 0.361 \\
\midrule
\multicolumn{2}{c}{\textbf{Mean $\pm$ Std}} &
$0.704 \pm .032$ & $0.625 \pm .027$ &
$0.657 \pm .007$ & $0.295 \pm .019$ & $0.354 \pm .030$ \\
\bottomrule
\end{tabular}%
}
\end{table}

\subsection{Position-Wise Feature Ablation}
Position-wise feature masking was conducted across positions N1--N4
to assess which encoded nucleotide positions contribute genuine
discriminative signal. Each position was independently zeroed at
inference time and the resulting change in ROC-AUC relative to the
unmasked baseline was recorded. Masking position N4 (corresponding
to downstream flanking nucleotides at positions $+2$ and $+3$ of the
8-mer) produced the only performance degradation ($\Delta$ROC-AUC $=
-0.021$), identifying it as the primary carrier of discriminative
signal under the current encoding scheme. Masking positions N1 and N2
produced ROC-AUC improvements of $+0.067$ and $+0.029$ respectively,
suggesting these upstream positions introduce noise under the current
4-qubit pair-condensation encoding. These results warrant further
investigation with a full 8-qubit encoding of all motif positions.
The complete ablation profile is illustrated in
Fig.~\ref{fig:combined_results} (panel c).

\section{Comprehensive Validation Protocol}
Evaluating classification performance requires metrics
and controls beyond overall accuracy. To ensure
scientific validity, we introduce a multi-tiered stress-testing and
validation framework integrated into our pipeline across four key
dimensions: multi-dimensional performance criteria, statistical
significance isolation, ablation studies, and validation
checkpointing. We report balanced accuracy, ROC-AUC, average precision,
and DNMT3B F1-score to characterize overall discrimination
and class-specific performance. Furthermore, executing the entire VQC pipeline across a
minimum of five entirely distinct random seeds allows us to report
accumulated evaluations as $\mu \pm \sigma$, establishing a rigorous
statistical baseline that rules out performance anomalies stemming
from fortunate weight initializations.

To isolate the true predictive capacity of the quantum circuit, the
model is systematically evaluated against three distinct control
frameworks designed to detect data exploitation over genuine
biochemical pattern acquisition. This includes: (1) a majority-class
baseline to verify that the VQC substantially outperforms a trivial
default model; (2) a shuffled-label test to confirm that performance
drops to random chance ($\text{ROC-AUC} \approx 0.500$) when
target label-to-feature associations are randomized; and (3) a
random-feature baseline that replaces structured genomic encoding
vectors with random Gaussian noise inputs, which produced equivalent
degradation. Our pipeline confirmed a ROC-AUC of $0.500$ and
accuracy at chance levels under shuffled labels.

\section{Conclusion \& Future Work}
We developed a four-qubit variational quantum classifier
to classify local flanking sequence preferences of
DNMT3A and DNMT3B, using a custom NumPy exact-statevector
simulator for the revised evaluation. The revised test partition contains 200 DNMT3A-labeled
and 210 DNMT3B-labeled motifs, giving a majority-class
accuracy baseline and positive-class prevalence of
approximately 0.512. Across five initialization seeds, the revised evaluation
reports a ROC-AUC of $0.661 \pm 0.009$, with variability
expressed as sample standard deviation. These results offer a promising 
proof-of-concept for deploying compact quantum learning networks to 
resolve complex biological sequence patterns.

Future work will examine alternative nucleotide
encodings, parameter-matched neural baselines,
shuffled-label and random-feature controls, and
generalization across independent experimental datasets.
Further studies will assess circuit behavior under
device-noise models and physical quantum execution.

\section{Acknowledgements}
S. Kais acknowledges support from the Goodnight Distinguished Chair in Quantum Computing at North Carolina State University.

\bibliographystyle{IEEEtran}
\bibliography{reference}

@article{zhong2014molecular,
  title={Molecular mechanism of action of plant DRM de novo DNA methyltransferases},
  author={Zhong, Xuehua and Du, Jiamu and Hale, Christopher J and Gallego-Bartolome, Javier and Feng, Suhua and Vashisht, Ajay A and Chory, Joanne and Wohlschlegel, James A and Patel, Dinshaw J and Jacobsen, Steven E},
  journal={Cell},
  volume={157},
  number={5},
  pages={1050--1060},
  year={2014},
  publisher={Elsevier}
}

@article{okano1999dna,
  title={DNA methyltransferases Dnmt3a and Dnmt3b are essential for de novo methylation and mammalian development},
  author={Okano, Masaki and Bell, Daphne W and Haber, Daniel A and Li, En},
  journal={Cell},
  volume={99},
  number={3},
  pages={247--257},
  year={1999},
  publisher={Elsevier}
}

@inproceedings{lo2016deep,
  title={Deep learning architectures for DNA sequence classification},
  author={Lo Bosco, Giosu{\`e} and Di Gangi, Mattia Antonino},
  booktitle={International Workshop on Fuzzy Logic and Applications},
  pages={162--171},
  year={2016},
  organization={Springer}
}

@article{preskill2018quantum,
  title={Quantum computing in the NISQ era and beyond},
  author={Preskill, John},
  journal={Quantum},
  volume={2},
  pages={79},
  year={2018},
  publisher={Verein zur F{\"o}rderung des Open Access Publizierens in den Quantenwissenschaften}
}

@article{bharti2022noisy,
  title={Noisy intermediate-scale quantum algorithms},
  author={Bharti, Kishor and Cervera-Lierta, Alba and Kyaw, Thi Ha and Haug, Tobias and Alperin-Lea, Sumner and Anand, Abhinav and Degroote, Matthias and Heimonen, Hermanni and Kottmann, Jakob S and Menke, Tim and others},
  journal={Reviews of Modern Physics},
  volume={94},
  number={1},
  pages={015004},
  year={2022},
  publisher={APS}
}

@article{schuld2019quantum,
  title={Quantum machine learning in feature Hilbert spaces},
  author={Schuld, Maria and Killoran, Nathan},
  journal={Physical review letters},
  volume={122},
  number={4},
  pages={040504},
  year={2019},
  publisher={APS}
}

@article{havlivcek2019supervised,
  title={Supervised learning with quantum-enhanced feature spaces},
  author={Havl{\'\i}{\v{c}}ek, Vojt{\v{e}}ch and C{\'o}rcoles, Antonio D and Temme, Kristan and Harrow, Aram W and Kandala, Abhinav and Chow, Jerry M and Gambetta, Jay M},
  journal={Nature},
  volume={567},
  number={7747},
  pages={209--212},
  year={2019},
  publisher={Nature Publishing Group UK London}
}

@article{schuld2020circuit,
  title={Circuit-centric quantum classifiers},
  author={Schuld, Maria and Bocharov, Alex and Svore, Krysta M and Wiebe, Nathan},
  journal={Physical Review A},
  volume={101},
  number={3},
  pages={032308},
  year={2020},
  publisher={APS}
}

@article{farhi2018classification,
  title={Classification with quantum neural networks on near term processors},
  author={Farhi, Edward and Neven, Hartmut},
  journal={arXiv preprint arXiv:1802.06002},
  year={2018}
}

@misc{b10,
  author = {Abraham, H. and others},
  title  = {{Qiskit}: An Open-Source Framework for Quantum Computing},
  year   = {2019},
  note   = {Available: https://doi.org/10.5281/zenodo.2562110}
}

@article{chakraborty2023quantum,
  title={Quantum computing on nucleic acid research: approaching towards next-generation computing},
  author={Chakraborty, Chiranjib and Bhattacharya, Manojit and Dhama, Kuldeep and Lee, Sang-Soo},
  journal={Molecular Therapy Nucleic Acids},
  volume={33},
  pages={53--56},
  year={2023},
  publisher={Elsevier}
}

@article{panizza2025quantum,
  title={Quantum computing for biophysical and optimization problems},
  author={Panizza, Veronica and others},
  year={2025},
  publisher={Universit{\`a} degli studi di Trento}
}

@article{lloyd2018quantum,
  title={Quantum generative adversarial learning},
  author={Lloyd, Seth and Weedbrook, Christian},
  journal={Physical review letters},
  volume={121},
  number={4},
  pages={040502},
  year={2018},
  publisher={APS}
}

@article{torlai2018neural,
  title={Neural-network quantum state tomography},
  author={Torlai, Giacomo and Mazzola, Guglielmo and Carrasquilla, Juan and Troyer, Matthias and Melko, Roger and Carleo, Giuseppe},
  journal={Nature physics},
  volume={14},
  number={5},
  pages={447--450},
  year={2018},
  publisher={Nature Publishing Group UK London}
}

@article{du2020expressive,
  title={Expressive power of parametrized quantum circuits},
  author={Du, Yuxuan and Hsieh, Min-Hsiu and Liu, Tongliang and Tao, Dacheng},
  journal={Physical Review Research},
  volume={2},
  number={3},
  pages={033125},
  year={2020},
  publisher={APS}
}

@article{monard1999combined,
  title={Combined quantum mechanical/molecular mechanical methodologies applied to biomolecular systems},
  author={Monard, Gerald and Merz, Kenneth M},
  journal={Accounts of chemical research},
  volume={32},
  number={10},
  pages={904--911},
  year={1999},
  publisher={ACS Publications}
}

@article{huembeli2021characterizing,
  title={Characterizing the loss landscape of variational quantum circuits},
  author={Huembeli, Patrick and Dauphin, Alexandre},
  journal={Quantum Science \& Technology},
  volume={6},
  number={2},
  pages={025011},
  year={2021},
  publisher={IOP Publishing}
}

@article{gelman2024survey,
  title={A survey of methods for mitigating barren plateaus for parameterized quantum circuits},
  author={Gelman, Michelle},
  journal={arXiv preprint arXiv:2406.14285},
  year={2024}
}

@article{wierichs2022general,
  title={General parameter-shift rules for quantum gradients},
  author={Wierichs, David and Izaac, Josh and Wang, Cody and Lin, Cedric Yen-Yu},
  journal={Quantum},
  volume={6},
  pages={677},
  year={2022},
  publisher={Verein zur F{\"o}rderung des Open Access Publizierens in den Quantenwissenschaften}
}

@inproceedings{isik2025multimodal,
  title={Multimodal Quantum Vision Transformer for Enzyme Commission Classification from Biochemical Representations},
  author={Isik, Murat and Saggi, Mandeep Kaur and Gowher, Humaira and Kais, Sabre},
  booktitle={2025 IEEE International Conference on Quantum Artificial Intelligence (QAI)},
  pages={135--140},
  year={2025},
  organization={IEEE}
}

@article{singh2026quantum,
  title={Quantum machine learning for complex systems: paradigms, applications, and challenges},
  author={Singh, Vinit and Bhatia, Amandeep Singh and Saggi, Mandeep Kaur and Sajjan, Manas and Kais, Sabre},
  journal={Academia Quantum},
  volume={3},
  number={2},
  year={2026},
  publisher={Academia. edu Journals}
}

@inproceedings{saggi2024mqml,
  title={Mqml: Multi-omic quantum machine learning based cancer classification, biomarker identification in human lung adenocarcinoma},
  author={Saggi, Mandeep Kaur and Kais, Sabre},
  booktitle={2024 IEEE International Conference on Quantum Computing and Engineering (QCE)},
  volume={1},
  pages={1713--1720},
  year={2024},
  organization={IEEE}
}

@article{bhatia2023quantum,
  title={Quantum machine learning predicting ADME-Tox properties in drug discovery},
  author={Bhatia, Amandeep Singh and Saggi, Mandeep Kaur and Kais, Sabre},
  journal={Journal of Chemical Information and Modeling},
  volume={63},
  number={21},
  pages={6476--6486},
  year={2023},
  publisher={ACS Publications}
}

\end{document}